\documentclass[10pt]{article}

\usepackage[margin=1in]{geometry}
\usepackage{times}
\usepackage{graphicx}
\usepackage{booktabs}
\usepackage{amsmath}
\usepackage{array}
\usepackage{xcolor}
\usepackage[hidelinks]{hyperref}
\usepackage{caption}
\usepackage{enumitem}
\usepackage{authblk}

\graphicspath{{./}}
\title{Multi-source Collaborative Training and Class-aware Fusion\\
for Sim2Real Roadside LiDAR 3D Detection}

\newcommand{\teamname}{\textbf{Team Name:} luopuu}

\author[1]{Pu Luo\thanks{25171214094@stu.xidian.edu.cn}}
\author[1]{Cong Xu\thanks{25171214052@stu.xidian.edu.cn}}
\author[1]{Yumei Li\thanks{25171213938@stu.xidian.edu.cn}}
\author[1]{Kexin Zhang\thanks{kxzh@stu.xidian.edu.cn}}
\author[1]{Licheng Jiao\thanks{lchjiao@mail.xidian.edu.cn}}
\author[1]{Wenping Ma\thanks{wpma@mail.xidian.edu.cn}}
\author[1]{Lingling Li\thanks{llli@xidian.edu.cn}}

\affil[1]{Xidian University, Xi'an, China}

\date{}

\begin{document}
\maketitle
\begin{center}
\teamname
\end{center}
\vspace{0.5em}

\begin{abstract}
Bridging the simulation-to-reality gap in roadside LiDAR requires addressing several coupled discrepancies, including scene geometry, sampling density, return patterns, and pedestrian scale. This report presents a multi-source collaborative training and class-aware fusion framework for Sim2Real 3D detection. The method organizes digital-twin scans, diffusion-redrawn scans, density-stabilized scans, and pedestrian morphology-aligned samples into a unified training pool with complementary roles. Within a common DSVT detection formulation, source-specialized expert branches preserve those roles while optimizing for the same detection objective. At inference, a predefined class-aware fusion pathway integrates geometry-stable and calibration-aware branches for vehicles, sampling-complementary branches for trucks, and morphology-consistent evidence for pedestrians. A label-free point-cloud center blend then refines geometric localization. On the UrbanTwin V2X-Real hidden test set, the unified system achieves a combined score of 0.7421, with 3D mAP@0.5 of 0.4518 and a realism score of 0.8871. The results indicate that a stable, interpretable collaboration among data sources is more valuable than unconstrained aggregation of model outputs.
\end{abstract}


\section{Introduction}
Roadside LiDAR is an important sensing modality for vehicle-infrastructure cooperation~\cite{v2xreal}. However, real-world point clouds are costly to annotate and their access can be limited. Digital twins offer scalable point clouds with accurate labels~\cite{urbantwin_hifi,urbantwin_data}, but detectors trained directly on synthetic data often face a substantial domain gap on real scans. This gap is not a single artifact: scene geometry, sampling density, sensor return style, and object morphology may differ simultaneously.

We therefore formulate Sim2Real perception as a collaborative learning problem supported by multiple synthetic data roles rather than as a direct mapping from one synthetic distribution to a real distribution. The proposed framework first establishes a multi-source training pool. It then trains source-specialized experts under a unified DSVT detection formulation and combines their outputs through fixed, class-aware pathways. This design retains complementary evidence while avoiding the indiscriminate averaging of correlated predictions.

The contributions of this work are fourfold:
\begin{itemize}[leftmargin=1.4em,itemsep=2pt]
    \item A multi-source synthetic training pool that jointly addresses geometric fidelity, realistic sampling style, density stability, and pedestrian morphology alignment.
    \item A source-specialized expert design in which all branches share one detection objective but retain complementary domain-aware inductive biases.
    \item A class-aware fusion mechanism and a label-free point-cloud center blend for stable semantic and geometric integration.
    \item An evaluation of the unified system on a fixed public validation protocol and the hidden benchmark set.
\end{itemize}

\section{Problem Formulation and Overview}
The task combines synthetic-point-cloud realism and real-domain 3D detection. Detectors are trained with synthetic labels only, while public real data are used for validation. The benchmark score is defined as
\begin{equation}
S = 0.6\cdot \mathrm{norm}(\mathrm{mAP}_{3D@0.5}) + 0.4\cdot \mathrm{norm}(R),
\end{equation}
where $R$ aggregates point-cloud distribution metrics including Chamfer Distance, RBF-MMD, Earth Mover's Distance, and Fr\'echet Point-cloud Distance after server-side normalization.

Our approach follows a single system-level chain: \emph{data roles $\rightarrow$ specialized experts $\rightarrow$ class-aware fusion $\rightarrow$ geometric refinement}. Each source serves a fixed purpose within the same training system. The resulting experts are not treated as competing candidate models; instead, their complementary coverage is explicitly routed to the classes for which it is relevant.

\begin{figure*}[t]
    \centering
    \includegraphics[width=\textwidth]{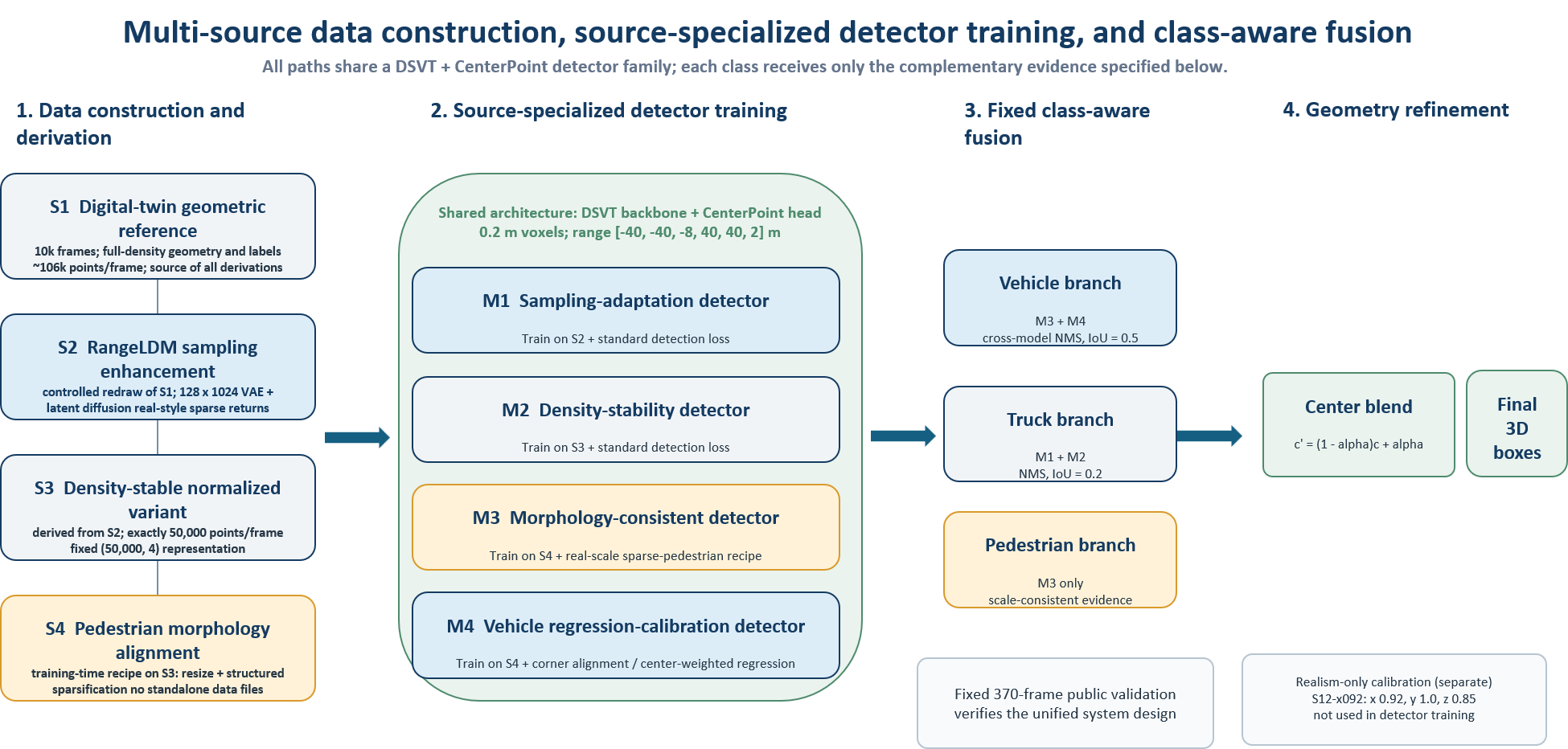}
    \caption{Multi-source collaborative training and class-aware fusion. Each data source has a fixed role within one unified detection system.}
    \label{fig:framework}
\end{figure*}

\section{Multi-source Collaborative Training}
\subsection{Multi-source Training Pool}
To prevent a single synthetic distribution from limiting detector generalization, we organize four sources as one unified data pool. Every source is constructed for the same three-class roadside LiDAR detection task (vehicle, pedestrian, and truck) and uses the same coordinate convention. The sources are not parallel, unrelated datasets: they form a derivation chain in which high-fidelity digital-twin geometry provides the reference, range-image diffusion changes the return pattern, density normalization controls the resulting input size, and a training-time morphology recipe adapts pedestrians. Consecutive scene fragments are split at the sequence level, with 250 frames per fragment, so temporally adjacent frames do not cross the training--validation boundary.

\begin{table}[t]
\centering
\small
\begin{tabular}{p{2.05cm}p{3.1cm}p{2.35cm}p{3.0cm}}
\toprule
\textbf{Data role} & \textbf{Construction} & \textbf{Scale / usage} & \textbf{Collaborative value}\\
\midrule
Geometry reference (S1) & High-fidelity UrbanTwin roadside digital twin & 10k frames & Full-density geometric and semantic reference; source of derived data\\
Sampling enhancement (S2) & RangeLDM controlled redraw of S1 & 2k frames & Real-style sparse return pattern; training input for Model~1\\
Density stabilization (S3) & Exact 50k-point resampling of S2 & 2k derived frames & Controlled density; training input for Model~2 and carrier for S4\\
Morphology alignment (S4) & Pedestrian resize and structured sparsification on S3 & Training-time recipe & Real-scale sparse pedestrians; training input for Models~3--4\\
\bottomrule
\end{tabular}
\caption{Roles of the multi-source training pool.}
\label{tab:sources}
\end{table}

\subsubsection{S1: geometric reference from a high-fidelity digital twin}
The starting point is a high-fidelity UrbanTwin roadside-intersection digital twin, constructed following the reproducible digital-twin workflow described by Shahbaz and Agarwal~\cite{urbantwin_hifi}. That workflow models static background geometry, road infrastructure, dynamic traffic, and sensor-specific specifications to create a synthetic environment aligned with a target roadside setting. Our S1 corpus contains 10,000 frames, with approximately 106k points per frame and 65.3k vehicle, 20.6k pedestrian, and 24.4k truck annotations in total. S1 preserves the full-density, undistorted geometric reference and all semantic labels. It is the foundation for S2 and, through the derived sources, for the complete detection system and the realism submission frames. The final four detection models are not trained directly on raw S1; its principal role is to provide a trustworthy geometric baseline and the source material for controlled derivations.

\subsubsection{S2: sampling enhancement through RangeLDM redraw}
S2 is created by applying RangeLDM~\cite{rangelm} to selected S1 frames. A VAE operates on $128\times1024$ range images and a latent diffusion process redraws the return pattern while retaining the scene's semantic structure. This allows the original labels to remain valid, with an observed point-retention rate of approximately 0.72 and labeled-object retention of approximately 0.97. Two non-overlapping redraw batches of 2,000 frames were generated to assess the reproducibility of the transformation; the final system fixes one 2,000-frame batch as its sampling-enhancement source. Compared with S1, S2 is sparser and better reflects the real sensor's sampling style. It is the training input for Model~1 and supplies one half of the complementary evidence used by the truck branch.

\subsubsection{S3: density-stable normalized variant}
Diffusion redraw changes not only the return pattern but also the number of observed points. To avoid input-size fluctuations becoming an unintended training variable, S3 is derived from S2 by resampling every frame to exactly 50,000 points with four attributes per point. The resulting 2,000-frame dataset has a fixed $(50000,4)$ representation per frame. S3 is used directly to train Model~2 and serves as the carrier data for the S4 training-time recipe. Thus, S2 contributes a sparse, realistic sampling style, whereas S3 contributes a controlled and stable density distribution; together they provide complementary coverage for long truck objects.

\subsubsection{S4: pedestrian morphology alignment and structured sparsification}
S4 is not stored as a separate dataset. Instead, it is a training-time recipe applied to S3. Each synthetic pedestrian is resized toward the real target distribution $[0.68,0.69,1.74]$ m. With probability 0.6, structured sparsification is then applied by removing one or two azimuth wedges and dropping 20--45\% of the remaining points at random. The motivation is a measured scale mismatch: real pedestrians have a median size of approximately $0.74\times0.72\times1.73$ m, whereas synthetic pedestrians are approximately $1.84\times0.80\times1.61$ m. The width discrepancy alone can limit the IoU of otherwise aligned boxes to about 0.35, below the 0.5 matching threshold. S4 is therefore the only source that contains both real-scale and sparse pedestrian evidence. It forms the data basis of Models~3 and~4 and raises pedestrian AP from zero to 0.142 under the local 50-frame diagnostic protocol.

\subsubsection{Realism-frame calibration outside the detection pool}
The realism task is calibrated separately from the detection data pool. The 50 submitted synthetic frames use affine candidate S12-x092, with $x\times0.92$, $y\times1.0$, and $z\times0.85$. This calibration is retained because it is stable under both local evaluation and the hidden reference, reaching an online realism score of 0.8871. Alternative transformations, including stronger $x$ scaling, quantile mapping, radial histogram matching, $z$-band reordering, and density resampling, produced overly optimistic local proxies but did not maintain hidden-reference performance. These alternatives are therefore not part of the final system.

\subsection{Source-specialized Collaborative Experts}
All experts use the DSVT sparse-voxel Transformer backbone~\cite{dsvt} and a CenterPoint-style detection head~\cite{centerpoint}, with a voxel size of 0.2 m and a perception range of $[-40,-40,-8,40,40,2]$ m. The branches share coordinate conventions, categories, and optimization objectives. Their distinction arises from the domain role emphasized during training rather than from incompatible architectures.

\begin{table}[t]
\centering
\small
\begin{tabular}{p{2.15cm}p{2.9cm}p{3.55cm}p{2.0cm}}
\toprule
\textbf{Expert role} & \textbf{Primary exposure} & \textbf{Collaborative responsibility}\\
\midrule
Sampling adaptation & Sampling enhancement & Learns real-style sparse sampling features\\
Density stability & Density stabilization & Maintains structural stability under dense input\\
Morphology consistency & Morphology alignment & Learns real-scale sparse pedestrians and vehicle structuren\\
Vehicle calibration & Morphology alignment with regression calibration & Supplements vehicle localization and corner constraints\\
\bottomrule
\end{tabular}
\caption{Source-specialized experts in the collaborative detection system.}
\label{tab:experts}
\end{table}

This organization converts data diversity into controlled model diversity. The sampling and density experts jointly cover large objects under different visibility patterns. The morphology-consistent expert protects pedestrian recognition from scale bias, while the vehicle calibration expert provides complementary localization constraints. The experts therefore form functional modules of one system rather than independent alternatives selected after the fact.

\subsection{Class-aware Cooperative Fusion}
Fusion is predefined from the semantic role of each source. Vehicles use mutually supported predictions from the morphology-consistent and vehicle-calibration experts, with cross-expert NMS at IoU 0.5. Trucks combine the complementary coverage of the sampling-adaptation and density-stability experts with NMS at IoU 0.2. Pedestrians are handled by the morphology-consistent expert alone so that scale-inconsistent predictions are not introduced. All categories use a final NMS threshold of 0.2 and a confidence threshold of 0.03.

This class-aware scheme avoids universal box averaging. Since experts share a backbone and much of the scene prior, their errors can be correlated; averaging all candidates can increase false positives. Fixed pathways instead restrict each class to evidence that matches its scale, density, and sampling characteristics.

\subsection{Geometry-consistent Point-cloud Center Blend}
After semantic fusion, every candidate box is geometrically refined using the test point cloud. Let $c$ be the predicted box center and $Q$ the points in a slightly expanded candidate box. The refined center is
\begin{equation}
c'=(1-\alpha)c+\alpha\cdot\mathrm{centroid}(Q),
\label{eq:centerblend}
\end{equation}
where $\alpha=0.20$ for vehicles and trucks and $\alpha=0.25$ for pedestrians. The formulation preserves network semantics while incorporating local point support, thereby reducing center bias in sparse observations without using real labels.

\begin{figure}[t]
    \centering
    \includegraphics[width=\linewidth]{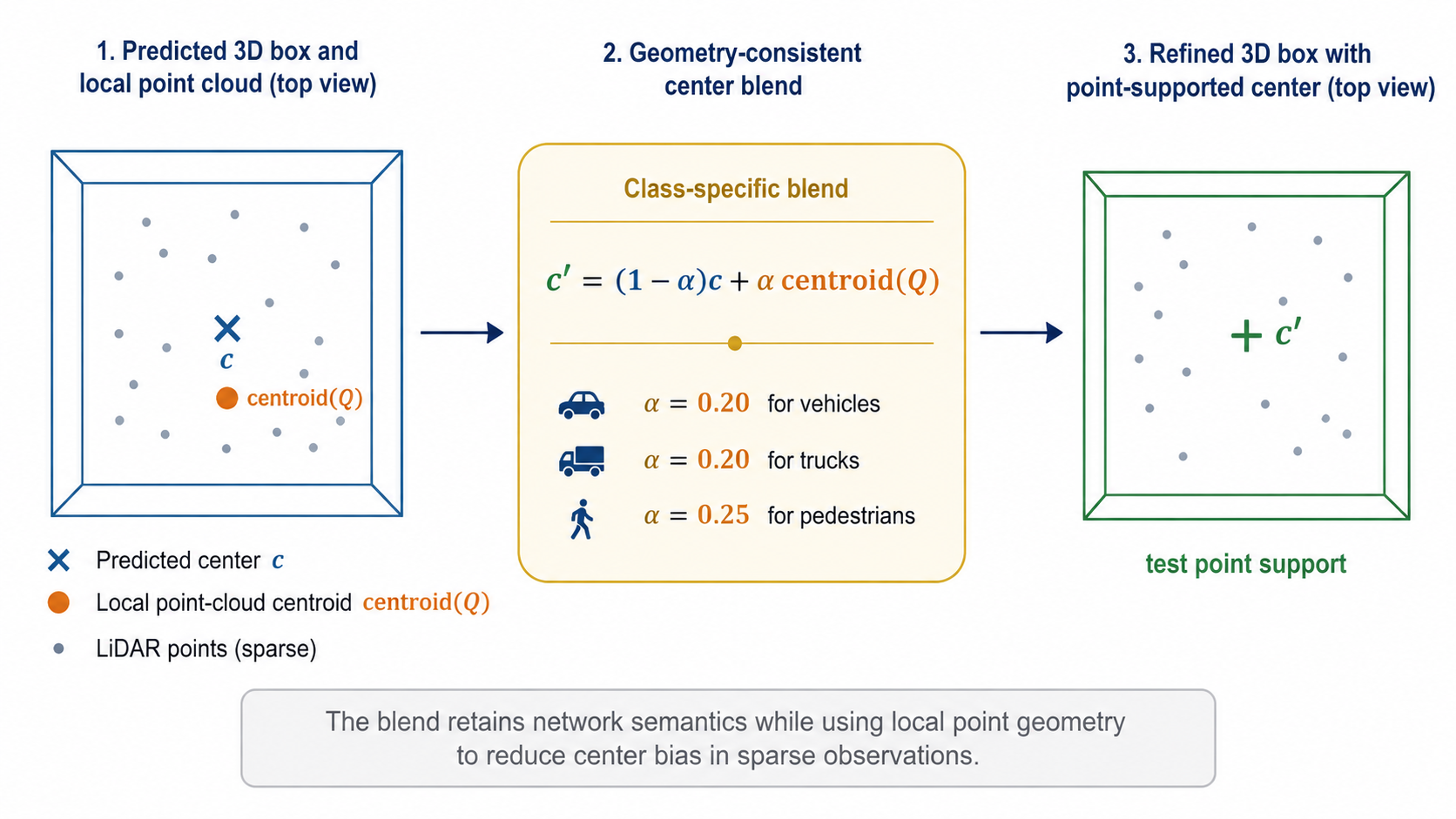}
    \caption{Geometry-consistent point-cloud center blend. The predicted center is refined using local point support without requiring real labels.}
    \label{fig:centerblend}
\end{figure}

\section{Experiments and Analysis}
\subsection{Experimental Setup}
The detection system is trained on the multi-source synthetic pool. A fixed 370-frame public validation protocol is used to verify the system design and to confirm the class-aware pathways. Training uses FP32, Dynamic VFE 128/128, and gradient clipping at 10. Source-specialized experts are trained for 40--60 epochs according to source scale. For realism, we use the geometry-preserving affine calibration S12-x092, corresponding to scale factors of 0.92, 1.0, and 0.85 along the $x$, $y$, and $z$ axes, respectively.

\subsection{Final Result}
Table~\ref{tab:final} reports only the final hidden-test result of the unified collaborative system. It achieves a combined score of 0.7421, 3D mAP@0.5 of 0.4518, a normalized detection score of 0.6455, and a normalized realism score of 0.8871.

\begin{table}[t]
\centering
\small
\begin{tabular}{lcccc}
\toprule
\textbf{Test set} & \textbf{Combined score} & \textbf{3D mAP@0.5} & \textbf{Detection norm.} & \textbf{Realism norm.}\\
\midrule
Hidden test set & 0.7421 & 0.4518 & 0.6455 & 0.8871\\
\bottomrule
\end{tabular}
\caption{Final result of the unified multi-source collaborative system.}
\label{tab:final}
\end{table}

The result reflects the joint effect of the training pool, class-aware fusion, and geometric refinement. The fixed validation protocol provides a stable check that the collaboration design generalizes beyond an individual source or a single class-specific effect.

\subsection{Realism and Fusion Analysis}
Realism calibration must preserve spatial geometry. Stronger axis-wise quantile transformations can improve local statistical proxies while degrading global structure measured by CD, EMD, and FPD. The selected affine calibration maintains a consistent geometric relationship and remains stable in benchmark evaluation.

Similarly, direct averaging among homogeneous experts does not provide a reliable gain, because their prediction errors are not independent. The class-aware pathways make the integration interpretable: data sources contribute according to their structural strengths, and each category receives only compatible evidence.

\section{Discussion}
The value of multi-source collaboration is not merely larger training volume or a larger number of models. Its value lies in assigning a clear role to every source in one system. Simply expanding similar diffusion data or relaxing box aggregation does not necessarily improve real-domain detection. In contrast, retaining complementary information about geometry, density, sampling, and object scale provides an interpretable basis for both training and fusion.

The approach depends on an initial diagnosis of the target-domain gap. If the sensor configuration, mounting height, or object distribution changes substantially, the source roles and fusion pathways should be revalidated. Future work may introduce uncertainty-aware adaptive fusion weights and incorporate sensor-physics priors into generative calibration.

\section{Conclusion}
This report presented a multi-source collaborative training and class-aware fusion framework for Sim2Real roadside LiDAR 3D detection. The method organizes digital-twin geometry, diffusion-based sampling enhancement, density stabilization, and pedestrian morphology alignment into a unified training pool. Source-specialized experts, fixed class-aware pathways, and label-free geometric refinement then ensure that these sources contribute coherently. The hidden-test result demonstrates the value of a system-level design centered on complementary data roles and geometric consistency rather than on post hoc aggregation of locally favorable outputs.

\bibliographystyle{plain}
\bibliography{Sim2Real_Roadside_LiDAR_Report}

@inproceedings{v2xreal,
  author    = {Xiang, Hao and Zheng, Zhaoliang and Xia, Xin and Xu, Runsheng and Gao, Letian and Zhou, Zewei and Han, Xu and Zheng, Xinhu and Bai, Lan and Yuan, Yunshuang and Zhao, Yingqian and Ma, Jiaqi},
  title     = {V2X-Real: A Large-Scale Dataset for Vehicle-to-Everything Cooperative Perception},
  booktitle = {European Conference on Computer Vision},
  year      = {2024}
}

@article{urbantwin_hifi,
  author  = {Shahbaz, Muhammad and Agarwal, Shaurya},
  title   = {PercepTwin: Modeling High-Fidelity Digital Twins for Sim2Real LiDAR-based Perception for Intelligent Transportation Systems},
  journal = {arXiv preprint arXiv:2509.02903},
  year    = {2025}
}

@article{urbantwin_data,
  author  = {Shahbaz, Muhammad and Agarwal, Siddharth},
  title   = {UrbanTwin: Synthetic Roadside LiDAR Datasets},
  journal = {IEEE Open Journal of Intelligent Transportation Systems},
  volume  = {7},
  pages   = {353--364},
  year    = {2026}
}

@inproceedings{rangelm,
  author    = {Hu, Qianjiang and Zhang, Zhimin and Hu, Wei},
  title     = {RangeLDM: Fast Realistic LiDAR Point Cloud Generation},
  booktitle = {European Conference on Computer Vision},
  year      = {2024}
}

@inproceedings{dsvt,
  author    = {Wang, Haiyang and Shi, Chen and Shi, Shaoshuai and Lei, Meng and Wang, Sen and He, Di and Schiele, Bernt and Wang, Liwei},
  title     = {DSVT: Dynamic Sparse Voxel Transformer with Rotated Sets},
  booktitle = {IEEE/CVF Conference on Computer Vision and Pattern Recognition},
  pages     = {13520--13529},
  year      = {2023}
}

@inproceedings{centerpoint,
  author    = {Yin, Tianwei and Zhou, Xingyi and Kr{\"a}henb{\"u}hl, Philipp},
  title     = {Center-Based 3D Object Detection and Tracking},
  booktitle = {IEEE/CVF Conference on Computer Vision and Pattern Recognition},
  pages     = {11784--11793},
  year      = {2021}
}

\end{document}